\pdfoutput=1

\documentclass[11pt]{article}

\usepackage{acl}

\usepackage{amsmath, amssymb, algorithm, algorithmic}

\usepackage{times}
\usepackage{latexsym}
\usepackage{CJKutf8}
\usepackage{graphicx}
\usepackage{changepage} 
\usepackage{xcolor,colortbl}
\usepackage{booktabs,siunitx}
\definecolor{indomain}{RGB}{144,238,144} 
\definecolor{inndomain}{RGB}{253,127,57} 
\definecolor{outdomain}{RGB}{173,216,230} 
\definecolor{lightblue}{rgb}{0.4, 0.7, 0.85} 
\definecolor{lightgray}{gray}{0.9}
\usepackage{tcolorbox}
\usepackage{mathrsfs}
\usepackage{array}
\usepackage{tabularx}
\usepackage{multirow}
\usepackage{verbatim}
\usepackage{amsmath}
\usepackage{amssymb}
\usepackage{arydshln}
\usepackage{verbatim}
\usepackage[normalem]{ulem}
\usepackage{rotating}
\usepackage{caption}
\usepackage{subcaption}
\usepackage{supertabular}
\usepackage{dsfont}
\usepackage{geometry}

\usepackage[utf8]{inputenc} % Allow UTF-8 input
\usepackage[T1]{fontenc}   

\usepackage{url}
\usepackage{xspace}
\usepackage{verbatim}
\usepackage{float}
\usepackage{multicol}

\usepackage{booktabs}  
\usepackage{makecell}
\usepackage{tabu}
\usepackage{tabulary}

\usepackage{amsfonts}    
\usepackage{mathtools}
\usepackage{nicefrac} 
\usepackage{tikz}

\usepackage{microtype}      % Microtypography
\usepackage{soul}
\usepackage{paralist}
\usepackage{pifont}
\usepackage{colortbl}

\usepackage{tcolorbox}
\newcommand{\CRAG}{\textsc{C-RAG}\xspace}

\usepackage{tcolorbox}

\title{Eliciting Critical Reasoning in Retrieval-Augmented Language Models via Contrastive Explanations}

\author{\textbf{Leonardo Ranaldi $^{(\dagger)}$ Marco Valentino $^{(\dagger)}$  Andr\'e Freitas$^{(\dagger,*, \ddagger)}$} \\
	$^{\dagger}$ Idiap Research Institute, Martigny, Switzerland \\
 $^{*}$Department of Computer Science, University of Manchester, UK \\
 $^{\ddagger}$National Biomarker Centre (NBC), CRUK Manchester Institute, UK \\
		{
  {\tt [name].[surname]@idiap.ch}
  } }

\begin{document}
\maketitle
\begin{abstract}
Retrieval-augmented generation (RAG) has emerged as a critical mechanism in contemporary NLP to support Large Language Models (LLMs) in systematically accessing richer factual context. However, the integration of RAG mechanisms brings its inherent challenges, as LLMs need to deal with potentially noisy contexts. Recent studies have shown that LLMs still struggle to critically analyse RAG-based in-context information, a limitation that may lead to incorrect inferences and hallucinations. In this paper, we investigate how to elicit critical reasoning in RAG via \emph{contrastive explanations}. 
In particular, we propose \emph{Contrastive-RAG} (\CRAG), a framework that \textit{(i)} retrieves relevant documents given a query, \textit{(ii)} selects and exemplifies relevant passages, and \textit{(iii)} generates explanations that explicitly contrast the relevance of the passages to \textit{(iv)} support the final answer. 
We show the impact of \textsc{C-RAG} building contrastive reasoning demonstrations from LLMs to instruct smaller models for retrieval-augmented tasks. Extensive experiments demonstrate that \CRAG improves state-of-the-art RAG models while \textit{(a)} requiring significantly fewer prompts and demonstrations and \textit{(b)} being robust to perturbations in the retrieved documents.

\end{abstract}

\begin{figure}[t]
\centering
    \includegraphics[width=\columnwidth]{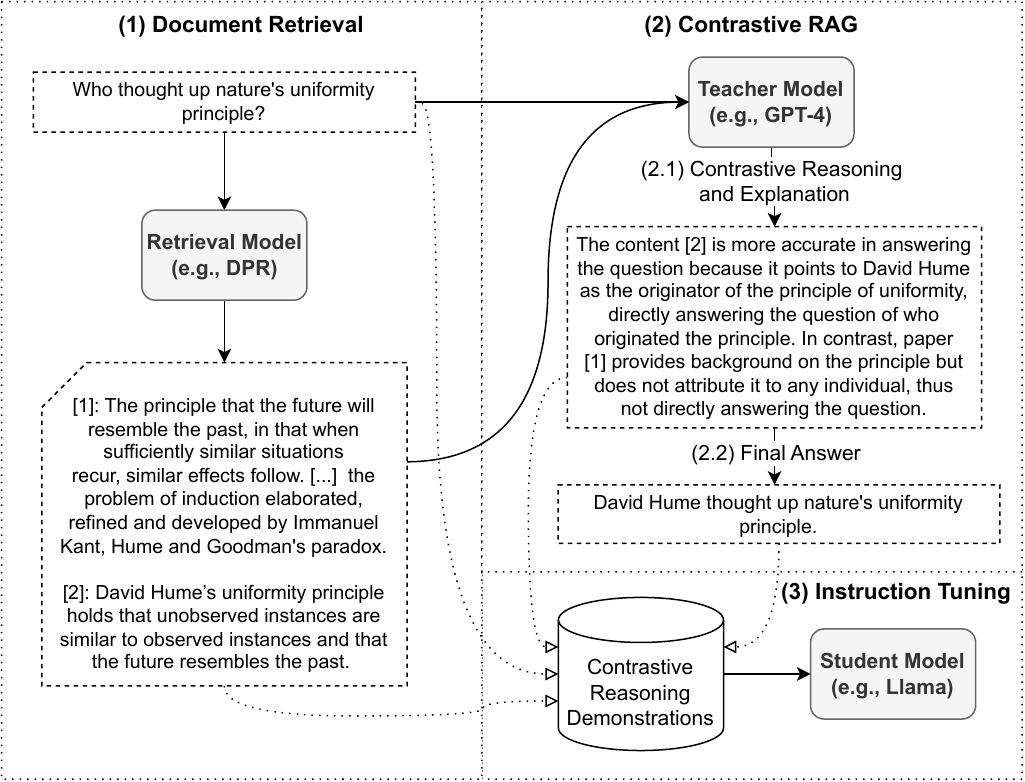}
    \caption{The overall pipeline of Contrastive-RAG (\S \ref{sec:methods}). \textit{(i)} A set of retried documents is provided as a context to an LLM that \textit{(ii)} generates contrastive explanatory arguments to arrive at the final answer. \textit{(iii)} The explanations generated by a teacher model are used as demonstrations to improve smaller student models.
    }
    \label{fig:pipeline_annotation}
\end{figure}

\begin{figure*}[h]
\centering
    \includegraphics[width=\textwidth]{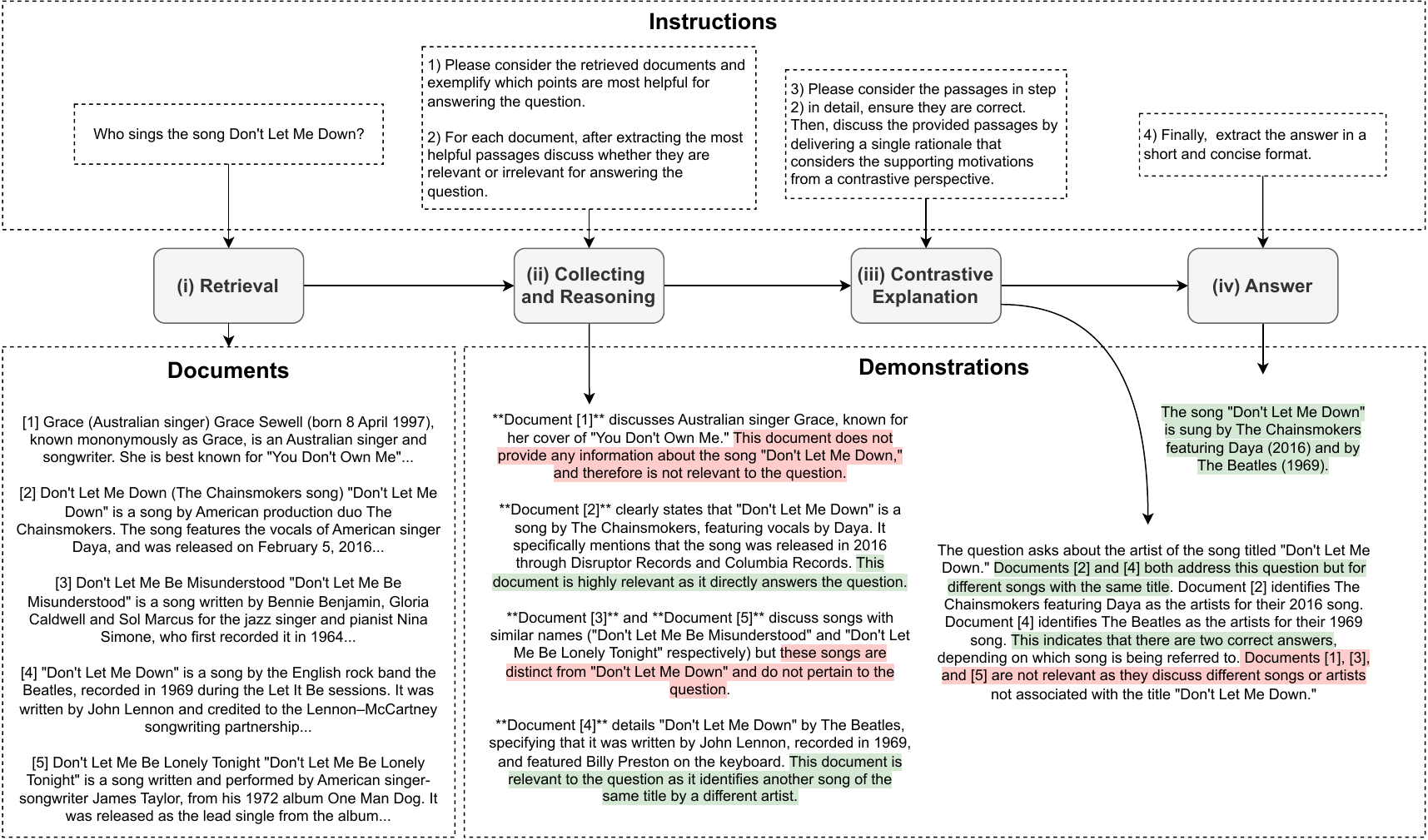}
    \caption{An illustration of the C-RAG framework for improving the traditional RAG pipeline. We demonstrate that \CRAG can significantly improve the performance of RAG models while requiring fewer prompts, training examples, and annotation steps than state-of-the-art approaches.}
    \label{fig:overall_pipeline}
\end{figure*}

\section{Introduction}
\label{sec:intro}
Retrieval-augmented generation (RAG) aims to improve the factuality and memory access of Large Language Models (LLMs) by systematically integrating relevant knowledge from external sources via retrieval mechanisms \cite{lewis2020retrieval}.
In particular, RAG is designed to mitigate some of the well-known limitations of LLMs, including the tendency for hallucinations and the lack of specific domain knowledge in the training data \cite{siriwardhana-etal-2023-improving,zhang2023sirenssongaiocean,kandpal2023largelanguagemodelsstruggle}.

Despite the benefits of RAG, contemporary studies have identified a range of persisting limitations emerging from noisy retrieval, where irrelevant or contradictory in-context passages can introduce biases in the models, particularly on smaller LMs \cite{petroni2020contextaffectslanguagemodels,shi2023largelanguagemodelseasily}. These shortcomings stem from the inability of RAG to systematically and critically assess the retrieved passages provided as context. While an emerging line of research attempted to improve the RAG pipeline by incorporating multi-step reasoning strategies to determine the relevance of in-context passages \cite{li-etal-2023-large,yoran2024makingretrievalaugmentedlanguagemodels}, this usually comes at the cost of significantly increasing the computational resources required for training and inference \cite{xia2024improvingretrievalaugmentedlanguage}.

To tackle such limitations, this paper proposes \emph{Contrastive-RAG} (\CRAG), a novel end-to-end framework designed to elicit a critical reasoning process in retrieval-augmented language models in the form of \emph{contrastive explanations} -- i.e., explanations that explicitly compare and contrast the relevance of the retrieved passages with respect to the task to be addressed. Specifically, the aim of \CRAG is to improve the original RAG pipeline via a multi-step reasoning framework composed of the following steps: \textit{i)} \textit{collecting step}, where query and documents are analysed to extract passages that are relevant for the query; \textit{ii)} \textit{contrastive reasoning}, where the LLMs construct explanatory arguments about the relevance of the extracted passages, highlighting and contrasting \textit{relevant} and \textit{irrelevant} knowledge; \textit{iii)} \textit{explanation}, where the contrastive arguments are consolidated into a single final explanation; and \textit{iv)} \textit{answering}, where a short-form answer is generated to address the query.

We demonstrate the impact of \CRAG by building reasoning demonstrations from LLMs and instructing smaller models to address retrieval-augmented tasks via contrastive explanations. 
An extensive empirical evaluation on four public question-answering (QA) tasks led to the following findings and conclusions: 
\begin{enumerate}
    \item Employing \CRAG to generate contrastive reasoning demonstrations via large reasoning models can significantly improve the performance of smaller RAG models leading to an average increase in accuracy of 55.4\% over RAG and 7.2\% and 1.9\% over Self-RAG \cite{asai2023selfraglearningretrievegenerate} and Self-Reasoning \cite{xia2024improvingretrievalaugmentedlanguage} respectively.
\item We found that \CRAG is significantly more efficient than state-of-the-art RAG frameworks requiring fewer prompts, annotation steps, and training examples (i.e., 2k vs 190k required by Self-RAG) while achieving better performance.
\item We demonstrate that \CRAG is robust to perturbations that are typically challenging for traditional RAG models, including the random shuffling of the retrieved documents and random noise applied to in-context passages.

\end{enumerate}

To the best of our knowledge, this is the first work to investigate the impact of contrastive explanations on RAG and demonstrate how contrastive reasoning demonstrations can consistently boost the performance of smaller LLMs, equipping them with the ability to critically analyse external knowledge and generate contrastive explanatory reasoning for their predictions. 

\section{Contrastive Explanations}

Contrastive explanations have been identified as fundamental modes of explanation in epistemology, artificial intelligence, and cognitive science \cite{lipton1990contrastive,miller2019explanation}. A contrastive explanation is a type of inference aimed at answering why-questions of the form \emph{``Why P rather than Q?''}, where \emph{P} is the explanandum -- i.e., the \emph{fact} to be explained -- and \emph{Q} is a counterfactual event -- i.e., the \emph{foil}. The function of a contrastive explanation is to describe what \emph{makes the difference} between the occurrence of the \emph{fact} and the \emph{foil}.

In this paper, we aim to leverage the notion of contrastive explanation to elicit critical reasoning in RAG. Our intuition is that contrastive explanations are a particularly fitting systematic reasoning mechanism for determining the \emph{relevance} of documents and passages provided as a context within the RAG pipeline. 

Formally, given a query $q$ and a set of documents $D = \{d_1, d_2, \ldots, d_n\}$, we want to partition $D$ into a set of relevant documents $P_q \in D$, and a set of irrelevant documents, $Q_q \in D$. In particular, we want these sets to be constructed through the generation of a natural language explanation $E$ that describes the factors that contribute to the relevance of $P_q$ and the factors that contribute to the irrelevance of $Q_q$. In the context of RAG, therefore, $E$ aims to answer the question \emph{``Why is $P_q$ relevant to $q$ rather than $Q_q$?''}, where the set of relevant documents $P_q$ corresponds to the \emph{fact}, and the set of irrelevant documents $Q_q$ corresponds to the \emph{foil}.

\section{Contrastive-RAG}
\label{sec:methods}

To elicit contrastive explanations in RAG, we present a framework composed of four inference stages (Figure \ref{fig:overall_pipeline}): i) \textit{collecting step} (\S \ref{sec:collecting}), where, given a query, the retrieved documents are collected and an LLM-based model extracts relevant passages; ii) \textit{contrastive reasoning} (\S \ref{sec:reasoning}), where an LLM-based model generates critical and contrastive explanations about the relevance of the extracted passages, by exemplifying and contrasting \textit{relevant} and \textit{irrelevant} aspects; iii) \textit{explanation} (\S \ref{sec:explaining}), where the arguments constructed in \textit{ii)} are summarised into a single explanation; iv) \textit{answering} (\S \ref{sec:explaining}), where a final short answer to the query is generated. The prompt adopted for \CRAG is illustrated in Table \ref{tab:annotation_prompt_CR-RAG}.
In this paper, we are interested in employing \CRAG to generate synthetic reasoning demonstrations (\S \ref{sec:annotations}) to enhance RAG models (\S \ref{sec:tuning}) and to equip them with the capability of critically analysing the questions and the retrieved documents and generate a contrastive explanation to arrive at the answer (Figure \ref{fig:overall_pipeline}).

\subsection{Collecting Step}
\label{sec:collecting}
RAG models leverage retrieved knowledge to improve accuracy and reduce hallucinations in LLMs. Therefore, the first step in the proposed pipeline involves retrieving relevant documents from a given reference document base $\mathcal{D}$. In this paper, we use DPR \cite{karpukhin-etal-2020-dense} and Contriever \cite{izacard2021unsupervised} as the default retriever models $R$. 
Subsequently, we instruct the LLM to analyse the question and extract the \textit{most relevant} passages from the set of retrieved documents (i.e., "\textbf{\#Reference Evidence}").
We collect the retrieved documents and relevant passages for all the queries and refer to this step as $\alpha_1$.

\subsection{Contrastive Reasoning}
\label{sec:reasoning}

Recent work has shown that exemplifying passages from the retrieved documents and directly using them as in-context knowledge for RAG can improve LLMs' accuracy \cite{asai2023selfraglearningretrievegenerate}. However, each passage may still contain irrelevant or contradictory knowledge that can mislead the model. 
Hence, after collecting the passages from the top-$k$ documents (\S \ref{sec:collecting}), we instruct the LLM to generate contrastive explanatory arguments to identify and compare relevant and irrelevant points in the passages with respect to the question (\S \ref{sec:collecting}). We perform this step by setting out the instructions as reported in Table \ref{tab:annotation_prompt_CR-RAG}. Hence, we collect the outcomes by defining this phase as $\alpha_2$.

\subsection{Explanation \& Answer}
\label{sec:explaining}

Finally, we leverage the arguments in the previous steps to generate a final contrastive explanation to derive the answer. In particular, we instruct the LLM to explicitly consider the contrastive rationales and summarise the main points into a single explanation. We define this step as $\alpha_3$.
Subsequently, we instruct the model to generate the final answer following the pattern "\textbf{\#Answer:}". This final step is defined as $\alpha_4$.

\subsection{\CRAG Operability}
\label{sec:annotations}

\CRAG leverages a set of structured instructions to deliver multi-step explanations via contrastive reasoning. Thus, the operability of \CRAG is two-fold, as it can be employed as both a prompting strategy and a synthetic annotation technique. 

\subsubsection{\CRAG as a Prompting Strategy}

By operating through the instructions in Table \ref{tab:annotation_prompt_CR-RAG}, we adopt \CRAG to prompt GPT-4 \cite{openai2023gpt4}. Specifically, we instruct GPT-4 to extract the most crucial passages from the retrieved documents (\S \ref{sec:collecting}), explaining the relevant and irrelevant points to answer the given question by exemplifying the main passages (\S \ref{sec:reasoning}), provide a single exhaustive explanation that best describes the critical points (\S \ref{sec:explaining}); and finally, generate the final answer in a strict format, in order to have a more straightforward and less ambiguous downstream assessment. 
However, although the sequence of instructions is well-structured and defined, the ability to perform sequential and complex reasoning tasks is limited to larger LLMs (such as GPT-4, as discussed in the experiments). Therefore, to transfer contrastive reasoning to smaller models, we use \CRAG to generate synthetic annotations as training demonstrations.

\subsubsection{\CRAG as a Demonstration Strategy}
\label{sec:annotation_strategy}
Following recent work \cite{xia2024improvingretrievalaugmentedlanguage,asai2023selfraglearningretrievegenerate}, we instruct smaller models via demonstrations produced by high-performing LLMs that are capable of following structured instructions. 
In contrast to previous methods, however, our approach is based on a single prompt composed of a series of sequential instructions (Figure \ref{fig:overall_pipeline}). 
Although GPT-4 have demonstrated the ability to follow sequential instructions \cite{peng2023instructiontuninggpt4}, we cannot formally guarantee that the generated demonstrations are correct. Therefore, we follow the method proposed by \citet{xia2024improvingretrievalaugmentedlanguage}, which computes the citation precision for the considered documents as a proxy for the quality of the demonstrations. However, since \CRAG uses a different annotation mechanism, our heuristics firstly filter out the final correct answers through a strict, exact match; then, after the filtering (cutting off about 50\% of the demonstrations), it verifies that each retrieved document along the reference evidence has been considered.
The starting annotations consist of approximately 10,000 training samples delivered with GPT-4, which, after the first filtering strategy, are reduced to 4,500 and finally, through quality control, are reduced to 2,000 (see Appendix \ref{app:annotation}).

\subsection{Training}
\label{sec:tuning}

Thanks to the operability of \CRAG (\ref{sec:annotations}), a Language Model $\pi$ can be trained using the generated annotations\footnote{we select annotations as described in Section \ref{sec:annotation_strategy}}, which are augmented with reasoning demonstrations $\alpha$ using the standard language modeling objective, maximizing likelihood:
\begin{equation}\label{eq:likehood}
\max _{\mathcal{\pi}} \mathbb{E}_{(q, \alpha, y) \sim \mathcal{D}_{A}} \log p_{\mathcal{\pi}}(y \mid \alpha, q) p_{\mathcal{\pi}}({\alpha \mid q})
\end{equation}

\noindent where $\alpha = \alpha_1 \oplus \alpha_2 \oplus \alpha_3 \oplus \alpha_4$ is the combination of the multiple reasoning steps performed by the model, $\oplus$ is the concatenation operator, and $\alpha_1$, $\alpha_2$, are the respective annotations generated by the above processes. $q$ is the provided question, and $y$ is the model output, including the intermediate steps and the final answer. $D_{A}$ is the training corpus augmented with contrastive reasoning demonstrations.

\section{Experiments}
\label{sec:exp_set}

We evaluate \CRAG on four open-domain question-answering tasks (\S \ref{sec:dataset}). We perform the retrieval and evaluation phases by following standard approaches used to assess the RAG pipeline (\S \ref{sec:retriever_evaluation}) and perform the tuning phase by using the setup presented in \S \ref{sec:training_setup}. 

\subsection{Tasks \& Datasets}
\label{sec:dataset}
We conduct an extensive experimental evaluation on the following question-answering (QA) tasks: (i) NaturalQuestion (NQ) \cite{kwiatkowski-etal-2019-natural}, (ii) PopQA \cite{mallen-etal-2023-trust}, (iii) TriviaQA \cite{joshi-etal-2017-triviaqa} and (iv) FEVER \cite{thorne-etal-2018-fever}. Appendix \ref{app:data_composition} describes the composition of each dataset.

\begin{table*}[t]
\small
\centering
\begin{tabular}{lcccccc}  
 \toprule 
 \multirow{1}{*}{\textbf{Models}} & \multicolumn{1}{c}{\textbf{NQ}} & \multicolumn{1}{c}{\textbf{PopQA}} & \multicolumn{1}{c}{\textbf{TriviaQA}} & \multicolumn{1}{c}{\textbf{FEVER}} & \multicolumn{2}{c}{\textbf{Train data for LLM}} \\  
 \cmidrule(lr){6-7}
\midrule
\multicolumn{5}{c}{\textbf{Baseline (no RAG)}} & \textbf{Training Size} & \textbf{Annotation} \\  
 \midrule 
Llama-2-7b & 19.2 & 18.4 & 30.5 & 20.1 & - & - \\  
Llama-2-13b & 24.0 & 22.6 & 38.5 & 25.2 & - & - \\
\hdashline 
\textsc{\textsc{C-RAG}-7b} & 30.0 & 44.8 & 58.6 & 32.5 & - & - \\ 
\textsc{\textsc{C-RAG}-13b} & 31.8 & 46.2 & 61.3 & 34.0 & - & - \\
\hdashline 
\rowcolor{gray!20} GPT-4-o & 35.2 & 52.4 & 64.3 & 36.5 & - & - \\  
\midrule
\multicolumn{5}{c}{\textbf{RAG}} & & \\  
 \midrule 
Llama-2-7b & 27.8 & 47.8 & 55.6 & 23.2 & - & - \\  
Llama-2-13b & 34.0 & 48.1 & 59.2 & 25.3 & - & - \\ 
\rowcolor{gray!20} GPT-4-o & 46.6 & 62.5 & 74.6 & 87.7 & - & - \\ 
\hdashline 
Llama-2-7b (\textbf{\textsc{C-RAG}}) & 24.6 & 47.0 & 54.8 & 22.9 & - & - \\ 
Llama-2-13b (\textbf{\textsc{C-RAG}}) & 33.5 & 47.4 & 58.0 & 24.9 & - & - \\
\rowcolor{gray!20} GPT-4-o (\textbf{\textsc{C-RAG}}) & 49.4 & 64.8 & 76.4 & 90.3 & - & - \\   
\midrule
\multicolumn{5}{c}{\textbf{RAG + Tuning (Llama-2-7b, -13b)}} & & \\  
 \midrule 
Llama-2-7b (SFT) & 36.8 & 54.4 & 61.9 & 67.5 & 2k & single-step \\
RECOMP 
& 38.4 & - & - & 39.1 & 150k & external \\
Self-RAG-7b  
& 37.2 & 54.9 & 66.4 & 70.2 & 190k & external \\
Self-RAG-13b 
& 38.8 & 55.8 & 67.2 & 72.2 & 190k & external \\
Self-Reasoning-7b 
& 38.0 & 54.2 & - & 78.6 & 2k & double-step \\
Self-Reasoning-13b 
& 41.4 & 57.3 & - & 83.9 & 2k & double-step \\
 \midrule 
\textsc{\textsc{C-RAG}-7b} & 40.2 & 56.4 & 68.4 & \textbf{79.2} & 2k & single-step \\  
\textsc{\textsc{C-RAG}-13b} & \textbf{42.6} & \textbf{58.2} & \textbf{70.3} & 83.6 & 2k & single-step \\  
\bottomrule
\end{tabular}
\caption{Overall results on QA and fact verification tasks \S \ref{sec:dataset}. The models are prompted as detailed in \S \ref{sec:prompting}, and the values correspond to the Exact Match (\%).
}
\label{tab:results}
\end{table*}

\subsection{Experimental Setup}
\label{sec:retriever_evaluation}

\paragraph{Retriever}
We use DPR \cite{karpukhin-etal-2020-dense} and Contriever-MS MARCO \cite{izacard2021unsupervised} to retrieve the top top-$k$ documents from the document base. We chose $k = 5$ to have a fair comparison to related RAG approaches using the same value for $k$. By default, we operate via DPR on NQ, as DPR has been fine-tuned on the dataset. On PopQA, where question and answer pairs are created based on Wikipedia, we use the Wikipedia\_corpus as background knowledge as proposed in \cite{xia2024improvingretrievalaugmentedlanguage}.

\paragraph{Evaluation Metrics}
We adopt two different evaluation metrics for QA and fact-verification tasks. Specifically with regard to QA tasks (NQ, PopQA, and TriviaQA), we use a flexible exact-match accuracy following \citet{schick2023toolformer,mallen-etal-2023-trust}, which is based on whether or not ground-truth answers are included in the generated answers provided by the models, instead of a strict exact match. For fact verification tasks, i.e., FEVER, we report the evaluation scheme proposed in \cite{thorne-etal-2018-fever} based on a three-class classification accuracy. Finally, \CRAG uses a special label \textbf{‘\#Answer’} (see Table \ref{tab:annotation_prompt_CR-RAG}) through which we instruct the models to deliver a short answer.

\subsection{Models Setup}
\label{sec:training_setup}

To get a comprehensive evaluation of existing RAG pipelines and the impact of \CRAG, three different LLMs are used: GPT-4 \cite{openai2023gpt4}, Llama-2-7 and Llama-2-13 \cite{touvron2023llama} along with their instruction-tuned chat version Llama-2-7-chat and Llama-2-13-chat. The models were chosen to have a common ground for comparison with state-of-the-art approaches.

\paragraph{Inference Settings}

We use greedy decoding in all experiments to ensure a more deterministic generation process. By using the prompt shown in Table \ref{tab:RAG_prompt} we set the temperature to 0.4 and maximum generation length of 2048, as we observed that these settings deliver better overall performance.

\subsubsection{Training Setup}
\label{sec:training_setting}

To evaluate the impact of \CRAG contrastive reasoning demonstrations on smaller models (\S \ref{sec:methods}), we use the annotations produced following the \CRAG strategy (\S \ref{sec:annotation_strategy}). Additionally, for a fair comparison, we produce annotations using Llama-2-SFT, where Llama-2 is fine-tuned on training samples without \CRAG. We compare our models with several related RAG approaches trained with demonstrations generated by GPT-4 to establish strong baselines (detailed in the last column of Table \ref{tab:results}). 
We fine-tune the Llama-2 models for 3 epochs with a batch size of 32 and a learning rate equal to 3e-5 with a 0.001 weight decay. We use the cosine learning rate scheduler with a warmup ratio of 0.03. We conducted our experiments on a workstation equipped with four Nvidia RTX A6000 with 48GB of VRAM.

\subsection{Prompting}
\label{sec:prompting}

We systematically prompt the models using two main settings:

\paragraph{Baseline (no RAG)}
We evaluate the baseline capabilities of selected models in a zero-shot manner and without introducing any in-context documents (without RAG) as in \cite{asai2023selfraglearningretrievegenerate,xia2024improvingretrievalaugmentedlanguage} using the prompt in Table \ref{tab:baseline_prompt}.

\paragraph{RAG Models}

We assess the impact of retrieved knowledge by instructing the evaluated models to consider the top-$5$ retrieved documents. In line with \cite{xia2024improvingretrievalaugmentedlanguage}, we use retrievers described in \S \ref{sec:retriever_evaluation} as in Table \ref{tab:RAG_prompt}. To complete the settings, we use \CRAG as a prompting strategy as in Table \ref{tab:annotation_prompt_CR-RAG}.

\begin{table}[t]
\small
\center
  \begin{tabularx}{0.46\textwidth}{p{1.8cm}<{\centering}p{1cm}<{\centering}p{1cm}<{\centering}p{1cm}<{\centering}}
    \toprule
    \multirow{2}{*}{ \textbf{Models} } & \multicolumn{1}{c}{\textbf{NQ}} & \multicolumn{1}{c}{\textbf{PQA}} & \multicolumn{1}{c}{\textbf{TQA}} \\
      \cmidrule(r){2-2}  \cmidrule(r){3-3}  \cmidrule(r){4-4} 
& (\%acc) & (\%acc) & (\%acc)  \\
    \midrule
\textsc{Full}  & \textbf{42.6} & \textbf{58.2} & \textbf{70.3} \\
\textsc{random}  & 38.4 & 54.2 & 61.0 \\

\midrule
    
w/o \textit{(2)}  & 33.2 & 52.2 & 60.6 \\
w/o \textit{(3)}  & 37.0 & 55.7 &  62.3 \\
w/o \textit{(4)}  & 40.8 & 57.4 &  69.4 \\
  \bottomrule
\end{tabularx}
\caption{Evaluation of impacts of each component on three QA tasks with \textsc{\textsc{C-RAG}-13b}. We eliminate (w/o) or \textsc{random} shuffling the four defined steps (\S \ref{sec:methods}). \textsc{\textsc{C-RAG}-7b}s' in Table \ref{tab:ablation_components_7}.}
  \label{tab:ablation_components}
\end{table}

\begin{figure*}[t]
\centering
         \begin{minipage}{0.24\linewidth}
     \centering
     \includegraphics[width=.95\linewidth]{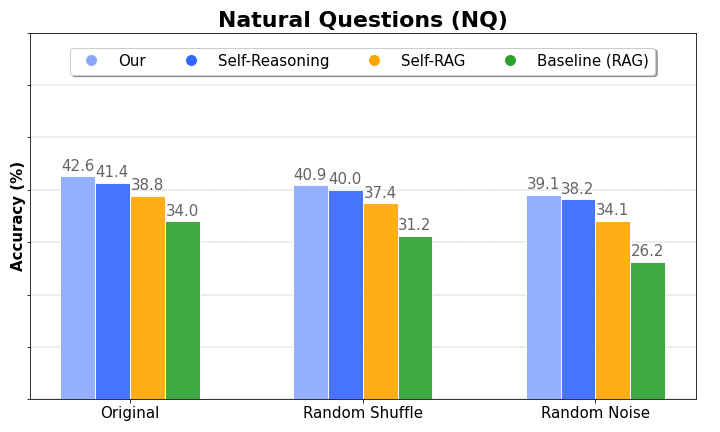}
   \end{minipage}
            \begin{minipage}{0.24\linewidth}
     \centering
     \includegraphics[width=.95\linewidth]{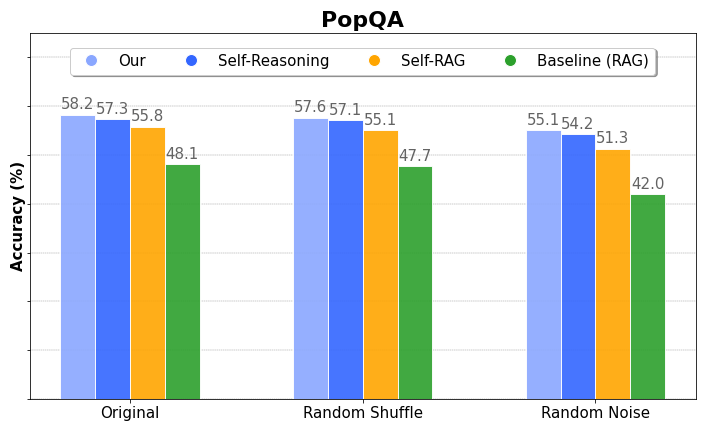}
   \end{minipage}
    \begin{minipage}{0.24\linewidth}
     \centering
     \includegraphics[width=.95\linewidth]{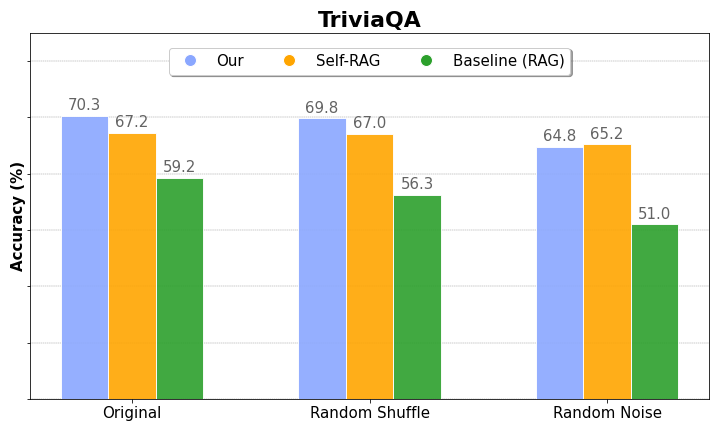}
   \end{minipage} 
       \begin{minipage}{0.24\linewidth}
     \centering
     \includegraphics[width=.95\linewidth]{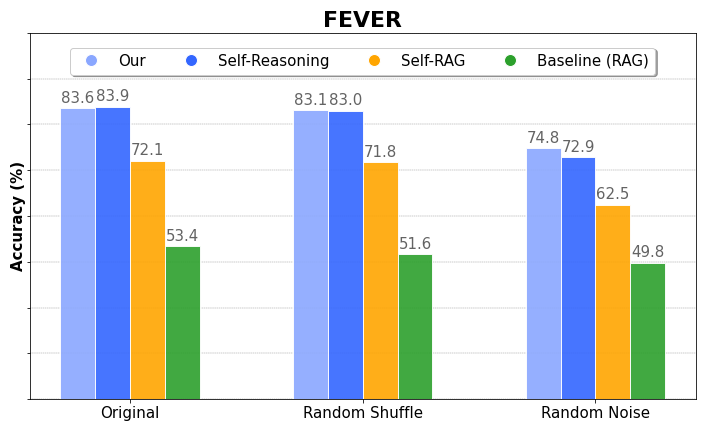}
   \end{minipage}

   \caption{Robustness experiment results on four QA datasets (\S \ref{sec:dataset}) using the same evaluation settings proposed for \textsc{\textsc{C-RAG}-13b} in Table \ref{tab:results}. We provide retrieved documents by randomly shuffling them (Random Shuffle) and using 50\% of irrelevant (Random Noise).} 
   \label{fig:performances_ablation_retrieving}

\end{figure*}

\section{Results \& Discussions}
\label{sec:Results}

The results of the empirical evaluation are reported in Table \ref{tab:results}. Overall, the experiments confirm that \CRAG can improve 
the capabilities of LLMs to handle retrieved documents for QA and fact verification tasks demonstrating the impact of contrastive reasoning and explanations on RAG models. We found that \CRAG is particularly effective as a demonstration strategy to improve the performance of smaller Llama-2 models, achieving state-of-the-art performance when compared to related fine-tuning approaches in the literature \cite{xu2023recompimprovingretrievalaugmentedlms,asai2023selfraglearningretrievegenerate,xia2024improvingretrievalaugmentedlanguage}. In the following sections, we analyse the impact of \CRAG when adopted as both a prompting strategy (\S \ref{sec:CR-RAG_in-context}) and as a framework for generating annotations to instruct LLMs (\S \ref{sec:CR-RAG_as_annotator}). 
Finally, we analyse the role of the contrastive explanations (\S \ref{sec:impact_explainations}) and provide evidence of robustness on challenging perturbations and low-resource settings (\S \ref{sec:robustness_ablation}).

\subsection{\CRAG as a Prompting Strategy}
\label{sec:CR-RAG_in-context}

The middle part of Table \ref{tab:results} reports the results of \CRAG when adopted as a prompting strategy for different models. While we can observe an overall improvement over the baseline models without RAG (with an improvement of 68.9\% for GPT-4, 56.6\% for Llama-2-7b and 65.4\% for Llama-2-13-b), the results show that the impact of \CRAG as a prompting strategy in a RAG setting is only evident for larger models (i.e., GPT-4-o) where \CRAG achieves an overall improvement of 3.9\%. For Llama-2-7b and Llama-2-13b, in fact, we observe a decrease in performance when compared to the standard RAG pipeline, indicating that such models are unable to generate the contrastive reasomning required to support their answers.

\subsection{\CRAG as a Demonstration Strategy}
\label{sec:CR-RAG_as_annotator}

The lower part of Table \ref{tab:results} reports the results of \CRAG when adopted as a demonstration strategy for different models. Here, the results show that \CRAG is highly effective in improving the performance of Llama-2 models when used to generate reasoning demonstrations via GPT-4. In particular, we found that \CRAG can outperform state-of-the-art approaches, including RECOMP (+1.8\%), Self-RAG (+7.2\%), and Self-Reasoning (+1.9\%). Moreover, as shown in Table \ref{tab:results}, \CRAG can achieve such results using a fraction of the reasoning demonstrations used by related approaches, also requiring fewer prompts and annotation steps. Specifically, our approach uses significantly fewer demonstrations when compared to RECOMP \cite{xu2023recompimprovingretrievalaugmentedlms} and Self-RAG \cite{asai2023selfraglearningretrievegenerate}, (2k demonstrations vs. 150k and 190k). Similarly, in contrast to Self-Reasoning \cite{xia2024improvingretrievalaugmentedlanguage}, \CRAG operates via a single-step prompting. This experimental setting significantly reduces the use of resources and simplifies the training process.

In addition, we observe that Llama-2 models fine-tuned via \CRAG can outperform GPT4-o by 6.1\% and substantially improve the performance of all the evaluated Llama-2 models (with and without RAG). These results strongly demonstrate the impact of the training signal provided by contrastive explanations and their ability to efficiently elicit critical arguments in smaller LMs (as shown in an inference example in Appendix \ref{app:example}).

\subsection{The Role of Contrastive Explanations}
\label{sec:impact_explainations}

Table \ref{tab:ablation_components} evaluates the impact of the end-to-end contrastive reasoning framework on the final performance. In particular, the table shows the effect of eliminating one of the steps or randomly shuffling them to produce the demonstrations.

The results demonstrate the importance of each stage in the multi-step reasoning process introduced in \S \ref{sec:methods}. In particular, we observe the highest decrease in performance when removing step 2 (i.e., Contrastive Reasoning, with a decrease of -17.2\%) and step 3 (i.e., Explanation, with a decrease of -10.4\%), demonstrating the crucial impact of the contrastive explanatory arguments on the final performance.

\subsection{Robustness \& Ablation Analysis}
\label{sec:robustness_ablation}

\begin{figure*}[t]
\centering
         \begin{minipage}{0.24\linewidth}
     \centering
     \includegraphics[width=.95\linewidth]{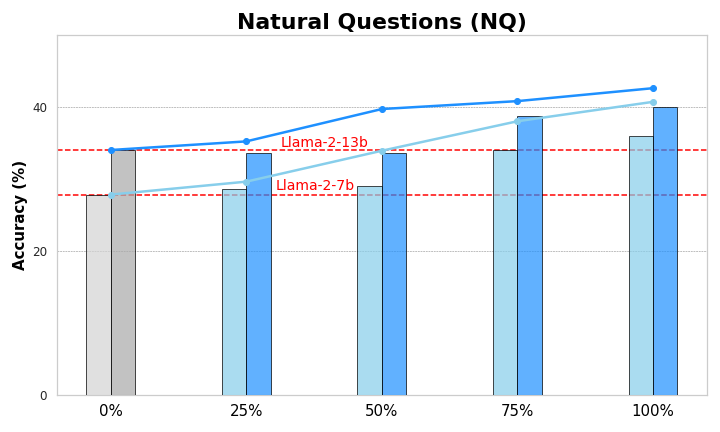}
   \end{minipage}
            \begin{minipage}{0.24\linewidth}
     \centering
     \includegraphics[width=.95\linewidth]{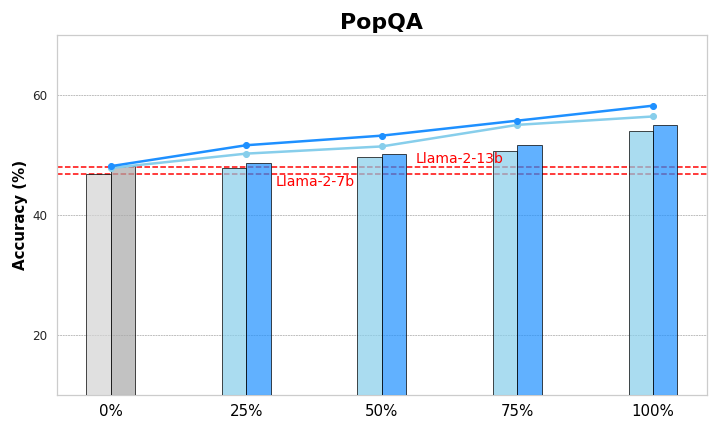}
   \end{minipage}
    \begin{minipage}{0.24\linewidth}
     \centering
     \includegraphics[width=.95\linewidth]{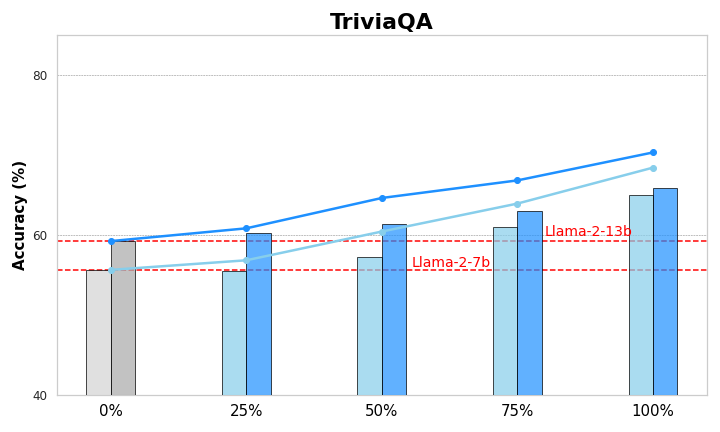}
   \end{minipage} 
       \begin{minipage}{0.24\linewidth}
     \centering
     \includegraphics[width=.95\linewidth]{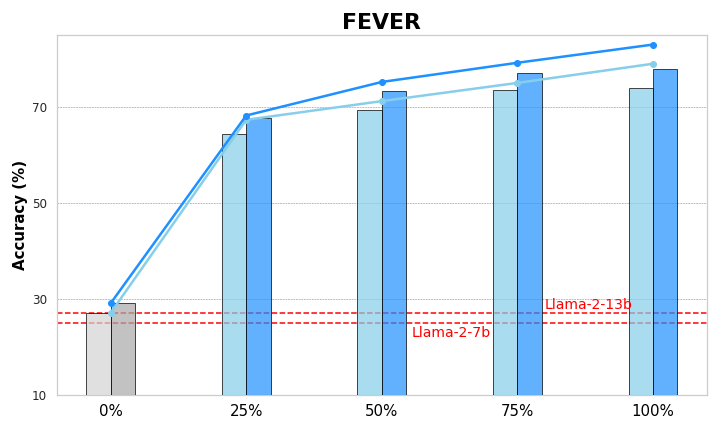}
   \end{minipage} 
          \begin{minipage}{.9\linewidth}
     \centering
     \includegraphics[width=.95\linewidth]{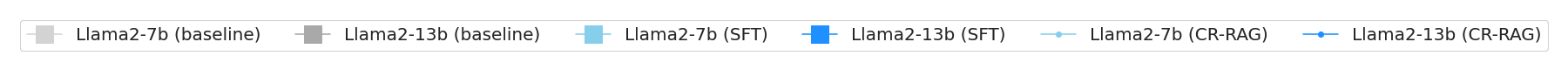}
   \end{minipage} 

   \caption{Performances assessment of \textsc{\textsc{C-RAG}-7b} and \textsc{-13b} by scaling training data. We replicated experimental settings proposed in \S \ref{sec:training_setup} changing the number of tuning instructions.} 
   \label{fig:performances_scaling_demons}

\end{figure*}

The \CRAG framework instructs models to reason on the retrieved documents independently of their order of occurrence (i.e., invariance to documents' permutations) and attempts to elicit critical explanations to identify irrelevant or contradictory knowledge in the extracted passages (i.e., robustness to the bias) as revealed in Figure \ref{fig:performances_ablation_retrieving}.
In addition, the benefits of \CRAG also emerge when reducing the number of demonstrations, showing that contrastive explanations can improve data efficiency (Figure \ref{fig:performances_scaling_demons}).
To assess such properties in more detail, we performed a robustness analysis along with an evaluation of how scaling the training demonstrations affects models' behaviours. 

\paragraph{Robustness to Perturbations}
Since noisy retrieval can negatively affect the performance of LLMs \cite{petroni2020contextaffectslanguagemodels}, we follow the methodology introduced in \cite{asai2023selfraglearningretrievegenerate,xia2024improvingretrievalaugmentedlanguage} to evaluate robustness. Specifically, we shuffled the order of the retrieved documents \textit{(Random Shuffle)} and inserted two irrelevant documents \textit{(Random Noise)}.
Figure \ref{fig:performances_ablation_retrieving} reports the experimental results. We found that the \CRAG framework consistently outperforms the baseline model (Llama-2-13b), as well as Self-RAG and Self-Reasoning, finding that perturbations have a lower impact on the final performance. In particular, the random shuffling of retrieved documents has a minimal impact on performance, demonstrating the permutation invariance property of \CRAG. 
Moreover, when noisy documents are added, all the evaluated models suffer a higher performance drop. However, the drop for \CRAG is typically lower than the RAG baseline, which shows that the proposed method is more robust even when dealing with noisier results. 

\paragraph{Quantity of Instructions} Figure \ref{fig:performances_scaling_demons} shows the behaviour of \CRAG when scaling-up the number of training examples. While we found that the quantity of the demonstrations used in \CRAG is important in determining the final performance, we found that \CRAG can outperform the baselines RAG models with only 50\% of training demonstrations, also achieving superior training performance when compared to the fine-tuned SFT model (i.e., the model fine-tuned without contrastive reasoning demonstrations as explained in \S \ref{sec:exp_set}).
This further highlights the quality of the training signal provided by the contrastive explanations.

\begin{table}[t]
\small
\centering
\setlength{\tabcolsep}{3pt} 
\begin{tabular}{l|c|cccc}
\textbf{Training} & \textbf{Model} & \multicolumn{4}{c}{\textbf{Evaluation}} \\
\cmidrule{3-6}
\multicolumn{2}{c|}{} & \textbf{NQ} & \textbf{PQA} & \textbf{TQA} & \textbf{FEV} \\
\toprule
& Baseline & 19.2 & 18.4 & 30.5 & 20.1 \\
& (+RAG) & 27.8 & 47.8 & 55.6 & 23.2 \\
\midrule
\textbf{NQ} & \texttt{SFT} & \cellcolor{inndomain} 36.5 & \cellcolor{indomain} 48.0 & \cellcolor{indomain} 56.2 & \cellcolor{indomain} 41.4 \\
& \textsc{C-RAG} & \cellcolor{inndomain} 39.8 & \cellcolor{indomain} 52.7 & \cellcolor{indomain} 62.6 & \cellcolor{indomain} 45.2 \\
\midrule
\textbf{PopQA}
 & \texttt{SFT} & \cellcolor{indomain} 31.0 & \cellcolor{inndomain} 50.2 & \cellcolor{indomain} 54.3 & \cellcolor{indomain} 40.4 \\
 &\textsc{C-RAG} & \cellcolor{indomain} 33.6 & \cellcolor{inndomain} 55.9 & \cellcolor{indomain} 57.4 & \cellcolor{indomain} 43.7 \\
\midrule
\textbf{TQA} & \texttt{SFT} & \cellcolor{indomain} 32.2 & \cellcolor{indomain} 49.6 & \cellcolor{inndomain} 60.7 & \cellcolor{indomain} 39.3 \\
 & \textsc{CRAG} & \cellcolor{indomain} 34.2 & \cellcolor{indomain} 52.2 & \cellcolor{inndomain} 68.0 & \cellcolor{indomain} 45.6 \\
\midrule
\textbf{FEVER} & \texttt{SFT} & \cellcolor{indomain} 29.4 & \cellcolor{indomain} 50.3 & \cellcolor{indomain} 52.9 & \cellcolor{inndomain} 66.6 \\
 & \textsc{C-RAG} & \cellcolor{indomain} 34.6 & \cellcolor{indomain} 53.0 & \cellcolor{indomain} 68.4 & \cellcolor{inndomain} 78.6 \\
\bottomrule
\end{tabular}
\caption{By replicating the experimental setting (\S \ref{sec:training_setup}), we trained the models (Llama-2-7b) on a single task and systematically evaluated them on other tasks. As \texttt{SFT}, we mean models (Llama-2-7b) instructed with input-output demonstrations consisting of queries, documents, and target answers.}
\label{tab:cross_eval_results}
\end{table}

\paragraph{Quality of Instructions} To complete the experiment, we assessed the transferability of \CRAG and analysed the impact of the quality of demonstrations using adversarial examples. Concerning transferability, we performed the training on a single task (we use the split reported in Table \ref{tab:data_composition}) and evaluated the models on another task. Table \ref{tab:cross_eval_results} shows that models trained through demonstrations can improve the performance of RAG models on tasks they have not been trained on.

Finally, we performed a sanity check to verify the impact of the quality of the demonstrations in Table \ref{tab:results_misleading_instructions} (in Appendix). Here, we performed adversarial experiments to study the impact of the quality of the demonstrations used to instruct the \CRAG models. 
We replicated the proposed experimental setting, operating via misleading instructions (defined in Appendix \ref{sec:misleading_training}) to ensure a complete understanding of the impact of the demonstrations.  
The results (Table \ref{tab:results_misleading_instructions}) show that the quality of the instructions matters. Models instructed with high-quality demonstrations (filtered as detailed in Appendix \ref{app:annotation}) achieve better performance than misleading demonstrations, which can degrade the accuracy below the baselines.

\section{Related Work}

Previous research investigated the advantages of augmenting Large Language Models (LLMs) through retrieved text passages, a technique known as Retrieval-augmented Generative (RAG) \cite{lewis2020retrieval,ram-etal-2023-context}.
However, recent work showed that the benefits of RAG can be undermined by noisy retrieval, thus decreasing consistency and reliability \cite{liu-etal-2023-evaluating,petroni2020contextaffectslanguagemodels,shi2023largelanguagemodelseasily}. 
Hence, several works proposed techniques to improve RAG reliability by enabling the models to elicit the critical steps to arrive at the final answer \cite{menick2022teachinglanguagemodelssupport,gao-etal-2023-enabling}. Similarly, recent work focused on improving the retrieval phase by fine-tuning LLMs to perform dynamic retrieval vi adaptive reasoning strategies \citet{jiang-etal-2023-active,yao2023reactsynergizingreasoningacting,gao-etal-2023-rarr,zhang2024raftadaptinglanguagemodel}. Nevertheless, the use of multi-step reasoning strategies usually comes at the cost of increasing the computational resources in terms of number of prompts and training examples\cite{10.1145/3637528.3671470,gao2024retrievalaugmentedgenerationlargelanguage}.

For instance, \citet{asai2023selfraglearningretrievegenerate} instructed models to retrieve information using special reflection tokens. However, this solution requires the training of two external models, requiring tens of thousands of additional training samples. \citet{xu2023recompimprovingretrievalaugmentedlms} attempted to lower the computational cost for multi-step RAG pipelines, but their approach still requires additional models to summarise the retrieved documents.
Finally, \citet{xia2024improvingretrievalaugmentedlanguage} eliminated the dependence on special tokens and external components by introducing reasoning trajectories that are employed to boost the performance of LLMs directly. Although the solution improves efficiency, the framework operates through a multi-step mechanism that requires multiple annotation phases. 

Similarly to recent work investigating the impact of natural language explanations on LLMs  \cite{he-etal-2024-using,dalal-etal-2024-inference,ye2022unreliability,quan2024verification,quan-etal-2024-enhancing,ranaldi-freitas-2024-aligning,ranaldi2024selfrefineinstructiontuningaligningreasoning}, we propose a method to integrate via multi-step explanations into RAG. To the best of our knowledge, however,  this is the first work to investigate the impact of contrastive explanations on RAG and demonstrate how contrastive reasoning demonstrations can boost the performance of smaller LMs.

\section{Conclusion}

RAG has shown great potential in boosting the performance of LLMs on knowledge-intensive tasks. Despite the success of RAG, noisy retrieval represents a major limitation. To tackle such challenges, we introduce a new framework called \CRAG, designed to improve RAG-based models by leveraging contrastive explanations.
We demonstrate that \CRAG can outperform state-of-the-art models while requiring fewer prompts and demonstrations and being robust to perturbations in the retrieved documents, laying the foundation for future research exploring the impact of natural language explanations on RAG-based models' efficiency, consistency and reliability.

\bibliography{anthology,custom}

\appendix

\newpage
\appendix

\begin{table*}

\section{Prompting Approaches}
\label{sec:app_prompting}

\begin{small}
\begin{tcolorbox}[colback=white, colframe=lightblue, sharp corners=south, rounded corners=north]

\begin{tcolorbox}[colback=white, colframe=black, rounded corners=south, rounded corners=north]
\textbf{\#Role} \\
You are an experienced expert skilled in answering various questions.
\end{tcolorbox}

\begin{tcolorbox}[colback=white, colframe=black, rounded corners=south, rounded corners=north]
\textbf{\#Task} \\
Please answer the question following the detailed requirements. 
\end{tcolorbox}

\begin{tcolorbox}[colback=white, colframe=black, rounded corners=south, rounded corners=north]
\textbf{\#Requirements} 

\begin{adjustwidth}{0.5cm}{0cm} 
    Please answer the question based on your knowledge using the format “\textbf{\#Answer:}” 
\end{adjustwidth}
\end{tcolorbox}

\begin{tcolorbox}[colback=white, colframe=black, rounded corners=south, rounded corners=north]
\textbf{\#Question} \\
\{\texttt{\textbf{question}}\}
\end{tcolorbox}

\end{tcolorbox}
\end{small}
\caption{Baseline prompting example.}
\label{tab:baseline_prompt}

\hspace{1cm}

\begin{small}
\begin{tcolorbox}[colback=white, colframe=lightblue, sharp corners=south, rounded corners=north]

\begin{tcolorbox}[colback=white, colframe=black, rounded corners=south, rounded corners=north]
\textbf{\#Role} \\
You are an experienced expert skilled in answering various questions.
\end{tcolorbox}

\begin{tcolorbox}[colback=white, colframe=black, rounded corners=south, rounded corners=north]
\textbf{\#Task} \\
Please answer the question based on the documents provided and following the detailed requirements using the format “\textbf{\#Answer:}” 
\end{tcolorbox}

% Reference Evidence
\begin{tcolorbox}[colback=white, colframe=black, rounded corners=south, rounded corners=north]
\textbf{\#Reference Documents} \\
 \hspace{1cm}  \textbf{[1]} \{\texttt{\textbf{Document}}$_1$\} \\
 \hspace{1cm}  \textbf{[2]} \{\texttt{\textbf{Document}}$_2$\} \\
  \hspace{1cm}  \textbf{[3]} \{\texttt{\textbf{Document}}$_3$\} \\
   \hspace{1cm}  \textbf{[4]} \{\texttt{\textbf{Document}}$_4$\} \\
  \hspace{1cm}  \textbf{[5]} \{\texttt{\textbf{Document}}$_5$\} 

\end{tcolorbox}

\begin{tcolorbox}[colback=white, colframe=black, rounded corners=south, rounded corners=north]
\textbf{\#Requirements} 

\begin{adjustwidth}{0.5cm}{0cm}

    Please consider the retrieved documents provided “\textbf{\#Reference Documents}” and answer the question. 
\end{adjustwidth}
\end{tcolorbox}

\begin{tcolorbox}[colback=white, colframe=black, rounded corners=south, rounded corners=north]
\textbf{\#Question} \\
\{\texttt{\textbf{question}}\}
\end{tcolorbox}

\end{tcolorbox}
\end{small}
\caption{Retrieval-augmented Generation prompting example.}
\label{tab:RAG_prompt}

\end{table*}

\begin{table*}
\section{\CRAG prompting Requirements}
\label{app:CR-RAG-prompt-annotation}

\begin{small}
\begin{tcolorbox}[colback=white, colframe=lightblue, sharp corners=south, rounded corners=north]

\begin{tcolorbox}[colback=white, colframe=black, rounded corners=south, rounded corners=north]
\textbf{\#Role} \\
You are an experienced expert skilled in answering various questions.
\end{tcolorbox}

\begin{tcolorbox}[colback=white, colframe=black, rounded corners=south, rounded corners=north]
\textbf{\#Task} \\
Please answer the question based on the documents provided and following the detailed requirements. 
\end{tcolorbox}

\begin{tcolorbox}[colback=white, colframe=black, rounded corners=south, rounded corners=north]
\textbf{\#Reference Documents} \\
 \hspace{1cm}  \textbf{[1]} \{\texttt{\textbf{Document}}$_1$\} \\
 \hspace{1cm}  \textbf{[2]} \{\texttt{\textbf{Document}}$_2$\} \\
  \hspace{1cm}  \textbf{[3]} \{\texttt{\textbf{Document}}$_3$\} \\
   \hspace{1cm}  \textbf{[4]} \{\texttt{\textbf{Document}}$_4$\} \\
  \hspace{1cm}  \textbf{[5]} \{\texttt{\textbf{Document}}$_5$\} 

\end{tcolorbox}

\begin{tcolorbox}[colback=white, colframe=black, rounded corners=south, rounded corners=north]
\textbf{\#Requirements} 

\begin{adjustwidth}{0.5cm}{0cm} 

    \textbf{1)} Please consider the retrieved documents provided “\textbf{\#Reference Documents}” and understand the main points. Follow the directions in detail and use only the information in the Documents, exemplifying which points are \textcolor{green}{most helpful} for answering the question \textbf{\#Question}.\\
    \textit{Do not forget any documents, and be as precise as possible.} \\ 
    
    \textbf{2)} For each document, after extracting the \textcolor{green}{most helpful} passages discuss whether they are actually \textcolor{blue}{relevant} or \textcolor{red}{irrelevant} for answering the \textbf{\#Question}. \\
    \textit{For clarity in your answer, provide the exact passages of each document referring to the document number, organizing the explanation as follows: passage claims [1] that [1], in contrast [4] claims.....}  \\ 

    \textbf{3)} Please consider the passages in step \textbf{2)} in detail, ensure they are correct. Then, discuss the provided passages by delivering a single rationale that considers the supporting motivations from a \textit{contrastive} perspective as concern \textcolor{blue}{relevant} and \textcolor{red}{irrelevant} passages. \\
    \textit{For clarity, provide a detailed explanation by marking it as “\textbf{\#Explanation:}”}
    \\

    \textbf{4)} Finally, after explaining the rationale supporting the final answer to facilitate the final evaluation, extract the answer in a short and concise format by marking it as “\textbf{\#Answer:}” 
\end{adjustwidth}
    
\end{tcolorbox}

% Question
\begin{tcolorbox}[colback=white, colframe=black, rounded corners=south, rounded corners=north]
\textbf{\#Question} \\
\{\texttt{\textbf{question}}\}
\end{tcolorbox}

\end{tcolorbox}
\end{small}
\caption{The Contrastive RAG (\CRAG) framework instructs the model to deliver multi-step reasoning paths that lead the models to solve the task by providing an explanation that contrasts the perspectives that have emerged.}
\label{tab:annotation_prompt_CR-RAG}

\end{table*}

\begin{table}[]
\section{Instructions Data}
\label{app:annotation}

As introduced in \S \ref{sec:methods}, we use our Contrastive RAG (\CRAG) to instruct Llama-2-7b and Llama-2-13 to address knowledge-intensive tasks using retrieved documents using \textit{contrastive} perspective (\S \ref{sec:tuning}). Since \CRAG alone do not benefit from the abilities of the baseline models (particularly Llama-2-7b and Llama-2-13b without further tuning as reported in Table \ref{tab:results}), we use GPT-4 (version \texttt{gpt-4o-2024-05-13}) as the annotation model. We systematically prompt GPT-4 using the instructions reported in Table \ref{tab:annotation_prompt_CR-RAG}.

We operate through GPT-4, which produced synthetic demonstrations, to instruct models to deliver \CRAG multi-step reasoned-solving strategies. Although this model addresses the tasks by following the instructions provided exhaustively \cite{peng2023instructiontuninggpt4}, these may still be incorrect and contain misleading information. Therefore, we checked the quality by filtering out high-quality demonstrations to refine the instruction set. 
Hence, by examining the answers, we eliminated all incorrect ones (i.e., all generations that do not contain the final target string metric better known as \textit{Exact Match}). We then checked that all the necessary steps were encoded in the remaining answers.

\end{table}

\begin{table}[h]
\section{Data Composition}
\label{app:data_composition}
\small
\centering
\setlength{\tabcolsep}{2pt}
\begin{tabularx}{0.4\textwidth}{lcccc}
    \toprule
    \textbf{Task} & \textbf{Total} & \textbf{Correct} & \textbf{\CRAG} & \textbf{Used} \\
    \midrule
    \textbf{NQ}        & $2,5k$ & $1,9k$ & $1.10k$ & $515$ \\
    \textbf{PQA}     & $2,5k$ & $1,1k$ & $0,75K$  & $500$ \\
    \textbf{TQA}      & $2,5k$ & $1,5k$ & $0,51K$   & $500$ \\
    \textbf{FEVER}     & $2,5k$ & $0,9k$ &  $0,48K$  & $485$ \\
    \hdashline
    \textbf{Total}     & $10k$   & $6,0k$ & $2,8k$ & \textbf{\textit{2,0k}} \\
    \bottomrule
\end{tabularx}
\caption{Data used to construct \CRAG instructions. We applied the annotation as explained in \S \ref{sec:annotation_strategy}. We obtained the following correct answers, filtered according to the heuristics in Appendix \ref{app:annotation}, and balanced for the four tasks. *($1k$ is equal to $1000$). }
\label{tab:data_composition}
\end{table}

\begin{table}[h]
\section{Eliminating Components}
We reproduced the experiment discussed in \S \ref{sec:Results} Table \ref{tab:ablation_components}. However, we used the -7b version unlike the previous one.

\small
\center
  \begin{tabularx}{0.46\textwidth}{p{1.8cm}<{\centering}p{1.2cm}<{\centering}p{1.2cm}<{\centering}p{1.5cm}<{\centering}}
    \toprule
    \multirow{2}{*}{ \textbf{Models} } & \multicolumn{1}{c}{ \textbf{NQ}} & \multicolumn{1}{c}{ \textbf{PopQA} } & \multicolumn{1}{c}{ \textbf{TriviaQA} } \\
      \cmidrule(r){2-2}  \cmidrule(r){3-3}  \cmidrule(r){4-4} 
& (\%acc) & (\%acc) & (\%acc)  \\
    \midrule
\textsc{Complete}  & \textbf{40.2} & \textbf{56.4} & \textbf{68.4} \\
\textsc{random}  & 32.0 & 52.6 & 57.0 \\

\hdashline
    
w/o \textit{(2)}  & 31.6 & 50.8 & 59.8 \\
w/o \textit{(3)}  & 36.8 & 54.2 &  61.3 \\
w/o \textit{(4)}  & 38.7 & 55.6 &  66.2 \\
  \bottomrule
\end{tabularx}
\caption{Ablation study on three QA task with \textsc{\textsc{C-RAG}-7b}. We analyze the impact of each component on tuning by eliminating (w/o) or \textsc{random} shuffling the four defined steps (\S \ref{sec:methods}).}
  \label{tab:ablation_components_7}
\end{table}

\begin{table}[t]
\section{Misleading \CRAG}
\label{sec:misleading_training}
Since the annotations produced through \CRAG are not always of good quality and correct (see Table \ref{tab:data_composition}
), we define these demonstrations as misleading (obtained through prompting \CRAG but with an incorrect final target). To observe their impact on tuning, we produced the experimental setting of \S \ref{sec:exp_set} by varying the demonstrations as shown in Table \ref{tab:results_misleading_instructions}. 

\small
\centering
  \begin{tabularx}{\columnwidth}{>{\hsize=1.6\hsize\arraybackslash}X 
                                  >{\hsize=0.8\hsize\centering\arraybackslash}X 
                                  >{\hsize=0.8\hsize\centering\arraybackslash}X 
                                  >{\hsize=0.8\hsize\centering\arraybackslash}X 
                                  >{\hsize=0.8\hsize\centering\arraybackslash}X}
    \toprule
    \textbf{Models} & \textbf{NQ} & \textbf{PopQA} & \textbf{TriviaQA} & \textbf{FEVER} \\
    \cmidrule(r){2-2} \cmidrule(r){3-3} \cmidrule(r){4-4} \cmidrule(r){5-5}
     & (\%acc) & (\%acc) & (\%acc) & (\%acc) \\
    \midrule
   \textit{baseline}  & 27.8 & 47.8 & 55.6 & 23.2 \\
    \textsc{CL-RAG-7b}  & \textbf{40.2} & \textbf{56.4} & \textbf{68.4} & \textbf{79.2} \\
\hdashline

    \textit{mixed}  & 28.0 & 49.8 & 56.7 & 67.3 \\
    \textit{misleading}  & 27.2 & 48.6 & 54.2 & 30.8 \\
    \bottomrule
  \end{tabularx}
\caption{The instructions used for our \textsc{CL-RAG-7b} and \textsc{-13b} are selected and filtred as detailed in \S \ref{sec:annotation_strategy}. We conducted tuning on incorrect (i.e., \textit{misleading}) and \textit{mixed} (50\% \textit{misleading} and 50\% filtred correct demonstrations) and reported the \textit{baseline} (Llama-2-7b).}
\label{tab:results_misleading_instructions}
\end{table}

\begin{table*}[h!]
\section{Example Generations}
\label{app:example}
\centering
\small
\begin{tabular}{p{16cm}}
\textbf{Question:} \textit{When was catch me if you can made?} \\
 \\
\midrule
\textbf{Llama-2-7b (baseline)} \\
The movie "Catch Me If You Can" is a biographical crime drama directed by Spielberg released in \textbf{2000}. \\
\midrule
\textbf{Llama-2-7b (RAG)} \\
Based on the information retrieved from Documents [1-5], we can determine that the Catch Me if You Can development began in 1980 and was released on \textbf{1989}. \\
\midrule
\textbf{Self-Reasoning-7b} (extracted from \cite{xia2024improvingretrievalaugmentedlanguage}) \\
\textit{Relevant:} True\\
\textit{Relevant Reason:} The provided documents are relevant with question. \\
\textit{Cite content:} [1] the original start date was January 2002, but was pushed to February 7 in
Los Angeles \\
\textit{Reason to cite:} This piece provides information on the commencement and location of filming for 'Catch Me If You Can', indicating that it started in April 2002 \\
\textit{Cite content:} [3] When production finished, the studio that had promised to distribute the movie had gone out of business, and it was eventually released by MCA Inc. in July 1989. \\
\textit{Reason to cite:} This piece provides information on the release date of 'Catch Me If You Can', indicating that it was eventually released by MCA Inc. in July. \\
The film 'Catch Me If You Can' was made in 2002. It started filming in April 2002 in Park Avenue, just outside the Waldorf-Astoria Hotel, and moved to Orange, New Jersey, before returning to Brooklyn for bank and courthouse scenes [1] \\
\midrule
\textbf{\textsc{\textsc{C-RAG}-7b}} \\
\uline{\# Reference Documents:}\\
- [1]: The film due to her busy schedule [...] \\
- [2]: Development for the film started in 1980 but did not book [...] \\
- [3]: The film was Sommers's directorial debut and was written by him. When production finished [...]\\

\textcolor{gray}{[other \# Reference Documents]} \\

\uline{\# Analysis:} \\

[1] \textit{Passage claims:} "The original start date was January 2002, but was pushed to February 7 in Los Angeles, California."
\textit{Relevance:} This is directly relevant as it provides the specific production start date for *Catch Me If You Can* in early 2002. \\

[2] \textit{Passage claims:} "Development for the film started in 1980, but did not progress until 1997, when Spielberg's DreamWorks bought the film rights to Abagnale's book."
\textit{Relevance:} Relevant for providing context on the film's development timeline, but no specific filming dates.
\\

[3] \textit{Passage claims:} "It was funded independently and had a budget of 800,000...released by MCA Inc. in July 1989."
\textit{Relevance:} \textbf{Irrelevant as it pertains to a different film.} \\

\uline{\# Explanation:} \\
\textit{Document \#1} provides the direct answer with filming starting in February 2002. \textit{Document \#2} gives development context but no dates for filming. \textit{Document \#3} is unrelated and irrelevant. \textcolor{gray}{(explanations for the remaining Docs)} \\

\uline{\# Answer:} \\
\textit{*Catch Me If You Can*} was made in early 2002, with filming beginning in February \textbf{2002}.
\\
\hline
\end{tabular}
\caption{An example of \CRAG inference compared to other approaches. \CRAG generates reasoned explanations, delivering the final correct answer. }

\end{table*}

\end{document}